\documentclass[conference]{IEEEtran}
\usepackage{cite}
\usepackage{amsmath,amssymb,amsfonts}
\usepackage{graphicx}
\usepackage{textcomp}
\usepackage{xcolor}

\usepackage{makecell}
\usepackage{multirow}
\usepackage{booktabs}
\begin{document}

\title{Patch-MoE Mamba: A Patch-Ordered Mixture-of-Experts State Space Architecture for Medical Image Segmentation}

\author{
\IEEEauthorblockN{
Diego Adame,
Fabian Vazquez,
Jose A. Nu\~nez,
Huimin Li,
Jinghao Yang,
Erik Enriquez
}

\IEEEauthorblockN{
DongChul Kim,
Haoteng Tang,
Bin Fu,
Pengfei Gu
}

\IEEEauthorblockA{
University of Texas Rio Grande Valley, Edinburg, TX 78539, USA\\
Email: pengfei.gu01@utrgv.edu
}
}

\maketitle

\begin{abstract}
CNN- and Transformer-based architectures have achieved strong performance in medical image segmentation, but CNNs are limited in modeling long-range dependencies, while Transformers often suffer from quadratic computational and memory complexity. State space models, especially Mamba-based networks, offer an efficient alternative with linear sequence complexity. However, existing Mamba segmentation models still face two limitations: pixel-wise directional scanning can disrupt local 2D spatial structure, and simple summation-based fusion of scan directions cannot adapt well to diverse object sizes, shapes, and boundaries. To address these issues, we propose \textit{Patch-MoE Mamba}, a patch-ordered mixture-of-experts state space architecture for medical image segmentation. It introduces a hierarchical patch-ordered scanning mechanism that preserves local spatial neighborhoods while capturing multi-scale context, and an MoE-based directional fusion module that adaptively combines multiple Mamba scanner outputs using four directional experts, a learnable concatenation expert, and residual directional aggregation. Experiments on five public polyp segmentation benchmarks and the ISIC 2017/2018 skin lesion segmentation datasets demonstrate the effectiveness and generality of Patch-MoE Mamba.
\end{abstract}


\section{Introduction}
Medical image segmentation is a fundamental task in computer-assisted diagnosis and treatment planning~\cite{fan2020pranet,tajbakhsh2015automated,zhang2022keep,an2025sli2vol+}. Advanced convolutional neural networks (CNNs) and Transformer-based architectures have achieved strong performance in this task~\cite{ronneberger2015u,gu2025self,dong2021polyp,zhang2023point}. However, CNNs are inherently limited by local receptive fields and thus may struggle to capture long-range dependencies, while Transformers often require large-scale labeled data and incur high computational and memory costs.

Recently, state space models, especially Mamba-based architectures, have attracted increasing attention for visual recognition and medical image segmentation because they can model global dependencies with linear complexity in sequence length~\cite{gu2023mamba,liu2024vmamba}. Despite their promise, existing Mamba-based segmentation networks still face two important limitations. First, most image Mamba models convert a 2D feature map into a 1D sequence and perform pixel-wise scanning along fixed directions~\cite{gu2023mamba,liu2024vmamba,ruan2024vm,zhang2024vm}. Although this design enables sequential modeling, it can weaken local 2D spatial coherence: spatially adjacent pixels may become distant in the scan sequence, and consecutive sequence elements may not be close in the image plane. This is undesirable for dense prediction tasks, where accurate boundary localization and local structure preservation are critical.

Second, current Mamba-based segmentation methods typically fuse directional scan outputs through simple summation~\cite{adame2025topo,ruan2024vm,zhang2024vm,vazquez2026learning}. Such fixed aggregation treats all directions and receptive fields as equally important at every spatial location. However, medical targets often vary substantially in size, shape, texture, and boundary complexity. Small or low-contrast lesions may require fine-scale local modeling, whereas larger or irregular structures may benefit from broader contextual information. Therefore, a hand-crafted summation strategy may limit the model's ability to adaptively emphasize the most informative directional responses.

To address these limitations, we propose \textit{Patch-MoE Mamba}, a patch-ordered mixture-of-experts (MoE) state space architecture for medical image segmentation. Instead of scanning individual pixels in a raster-like order, we partition feature maps into local patches and perform directional scanning in a patch-ordered manner. This strategy keeps pixels within each patch consecutive in the sequence, thereby better preserving local spatial neighborhoods while still enabling long-range dependency modeling across patches. We further employ hierarchical patch sizes at different stages to capture both fine-grained boundary details and coarse semantic structures.

In addition, we introduce an MoE-based directional fusion module to replace fixed summation. Specifically, the outputs of four directional Mamba scanners and their learnable concatenation are treated as five experts. A spatial-aware gating network computes location-dependent expert weights to adaptively fuse these directional responses. We also add a residual summation of the raw directional outputs to stabilize training and preserve strong directional signals. Experiments on five public polyp segmentation datasets and two skin lesion segmentation datasets show consistent improvements over CNN-, Transformer-, and Mamba-based baselines, demonstrating the effectiveness and generality of Patch-MoE Mamba.

\begin{figure*}[t]
    \centering
    \includegraphics[width=0.9\linewidth, keepaspectratio]{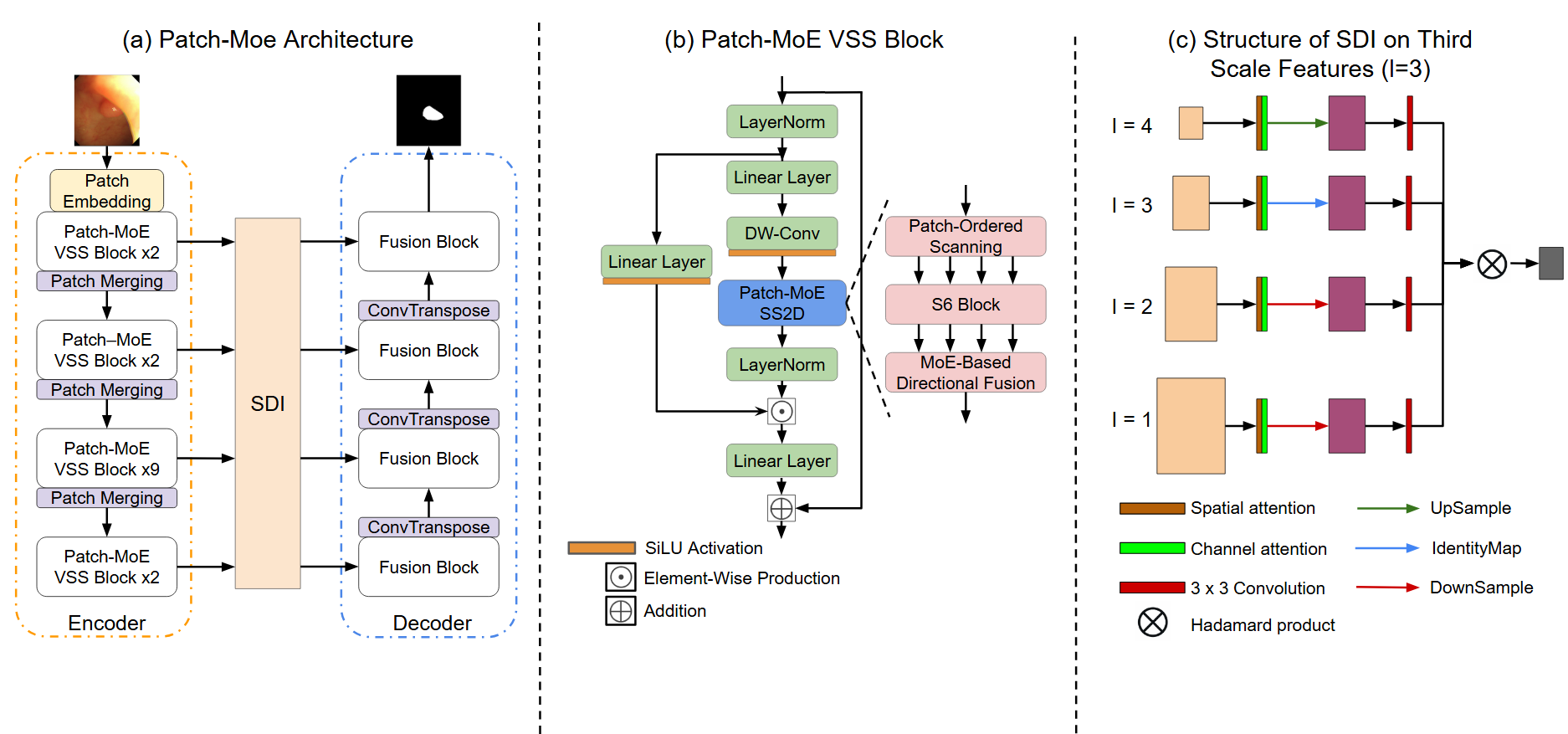}
    \caption{(a) Overview of the proposed Patch-MoE Mamba architecture. (b) Structure of the Patch-MoE Visual State Space (VSS) block. (c) Structure of the Semantics and Detail Infusion (SDI) module. For simplicity, only the refinement of features at the third scale ($l=3$) is shown.}
    \label{fig:pipeline}
\end{figure*}

%%%%%%%%%%%%%%%%%%%%%%%%%%%%%%%%%%%%%%
\section{Methods}
\subsection{Patch-MoE Mamba Architecture}
Fig.~\ref{fig:pipeline}(a) shows the overall architecture of Patch-MoE Mamba. It follows a U-Net-style design, consisting of a Mamba-based encoder, a semantics and detail infusion (SDI) module, and a decoder. Specifically, we replace the visual state space (VSS) block in VM-UNetV2~\cite{zhang2024vm} with a new Patch-MoE VSS block that integrates a patch-ordered scanning mechanism (Section~\ref{sec:patch-scan}) and an adaptive MoE fusion module (Section~\ref{sec:moe}) to form the encoder (Fig.~\ref{fig:pipeline}(b)). We then adopt the SDI module from U-Net v2~\cite{peng2025u} (Fig.~\ref{fig:pipeline}(c)) to enhance the feature maps at each level by infusing semantic information from higher-level features and integrating finer details from lower-level features via the Hadamard product. Finally, we retain the decoder design of VM-UNetV2~\cite{zhang2024vm}.

% Patch-MoE Mamba retains the overall UNet-like structure and SDI-based decoder of VM-UNetV2~\cite{zhang2024vm}, while replacing the original Vision Mamba encoder with a new architecture that integrates patch-ordered scanning and adaptive mixture-of-experts fusion. Our redesigned encoder addresses two limitations in existing Mamba-based segmentation: (1) pixel-level raster scanning disrupts spatial locality during sequence construction, and (2) fixed summation of directional scan outputs prevents context-dependent fusion.
\subsection{Patch-Ordered Scanning}
\label{sec:patch-scan}
\begin{figure}[t]
\centering
\includegraphics[width=0.9\linewidth]{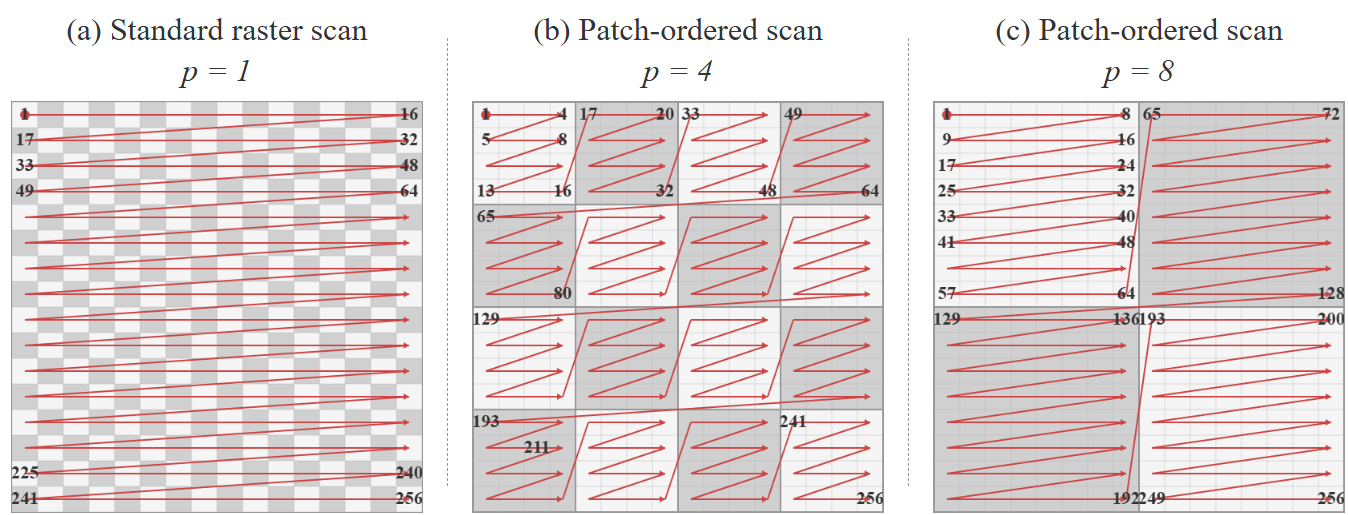}
\caption{Illustration of the patch-ordered scanning method on a $16 \times 16$ grid. 
%(a) Standard raster scan traverses pixels row by row, so vertically adjacent pixels (e.g., positions 1 and 17) are separated by 16 steps in the 1D sequence. 
%(b) With patch size $p=4$, pixels are grouped into non-overlapping $4 \times 4$ patches and scanned patch by patch; all 16 pixels within a patch are processed consecutively before moving to the next patch. 
%(c) With patch size $p=8$, larger $8 \times 8$ patches group 64 pixels each, further increasing spatial locality within contiguous segments of the sequence. 
For simplicity, only the forward scanning direction is shown.}
\label{fig:patch-scan}
\end{figure}

Existing Mamba-based models~\cite{liu2024vmamba,ruan2024vm,zhang2024vm} typically treat a 2D feature map as a long 1D token sequence and apply pixel-level raster scanning. As illustrated in Fig.~\ref{fig:patch-scan}(a), standard raster scanning traverses the grid row by row. While this formulation leverages the sequential modeling strength of state space models, it implicitly destroys local 2D structure: pixels that are adjacent in the scan order may be far apart spatially, and spatially neighboring pixels can be separated by a large sequence distance. For example, vertically adjacent pixels (positions 1 and 17 in the $16 \times 16$ grid) are separated by 16 steps in the 1D sequence. As a result, important local patterns such as lesion boundaries, fine surface structures, and local context around small or low-contrast objects are diluted or distorted in the sequence representation. This pixel-wise scanning thus tends to lose spatial coherence, which is undesirable for dense prediction tasks (e.g., image segmentation).

To preserve spatial coherence during sequence construction, we introduce a \emph{patch-ordered} scanning strategy in place of the raster-based flattening used by standard Vision Mamba modules. Given a feature map
$
X_l \in \mathbb{R}^{C_l \times H_l \times W_l}
$
and a patch size $p$, we partition the spatial grid into
$\lceil H_l / p \rceil \times \lceil W_l / p \rceil$ non-overlapping patches of at most $p \times p$ pixels (Fig.~\ref{fig:patch-scan}(b)). Within each patch, we enumerate all pixel locations $(r,c)$ in row-major order and append their indices to a global index list before proceeding to the next patch. This process yields a permutation vector of length $H_l \times W_l$ that reorders spatial positions but retains every pixel. In other words, the full-resolution feature grid is preserved and no pooling or token reduction occurs; only the visiting order changes. Compared to raster scanning, patch-ordered scanning ensures that all pixels inside a patch are mapped to consecutive positions in the sequence, thereby improving local spatial coherence while still allowing the state space model to capture long-range dependencies across patches.

To increase spatial locality within contiguous segments of the sequence and to encode multi-scale context, we develop a hierarchical patch-based scanning mechanism (Fig.~\ref{fig:patch-scan}(c)). Instead of using a single patch size, we define a set of patch sizes $\{p^{(1)}, p^{(2)}, \ldots\}$ that includes both small and large patches. Larger patches (e.g., $p^{(2)} = 8$) group more pixels into each contiguous sequence segment, strengthening the modeling of coarser structures and broader regions, while smaller patches (e.g., $p^{(1)} = 4$) enforce fine-grained locality and are well suited for capturing subtle boundary details. For each patch size $p^{(k)}$, we construct a corresponding patch-ordered index sequence as described above. These hierarchical sequences enable the Mamba blocks to model both fine- and coarse-scale spatial neighborhoods while preserving resolution.

Following VM-UNetV2~\cite{zhang2024vm}, we employ four scanning directions in each VSS block: (i) forward (left-to-right, top-to-bottom), (ii) reverse, (iii) width–height (WH) forward (top-to-bottom, left-to-right), and (iv) WH reverse. For each direction $d$, we apply patch-ordered scanning with a possibly distinct patch size $p^{(k_d)}$ drawn from the hierarchical set. This design enables each directional Mamba scanner to operate with its own spatial granularity—e.g., finer patches for horizontally oriented scans and coarser patches for vertically oriented scans—jointly capturing anisotropic and multi-scale spatial structures. The resulting directional sequences are then processed by the state space model and subsequently fused by our MoE-based fusion module (Section~\ref{sec:moe}).

%%%%%%%%%%%%%%%%%%%%%%%%%%%%%%%%%%%%%%
\begin{figure}[t]
\centering
\includegraphics[width=0.9\linewidth]{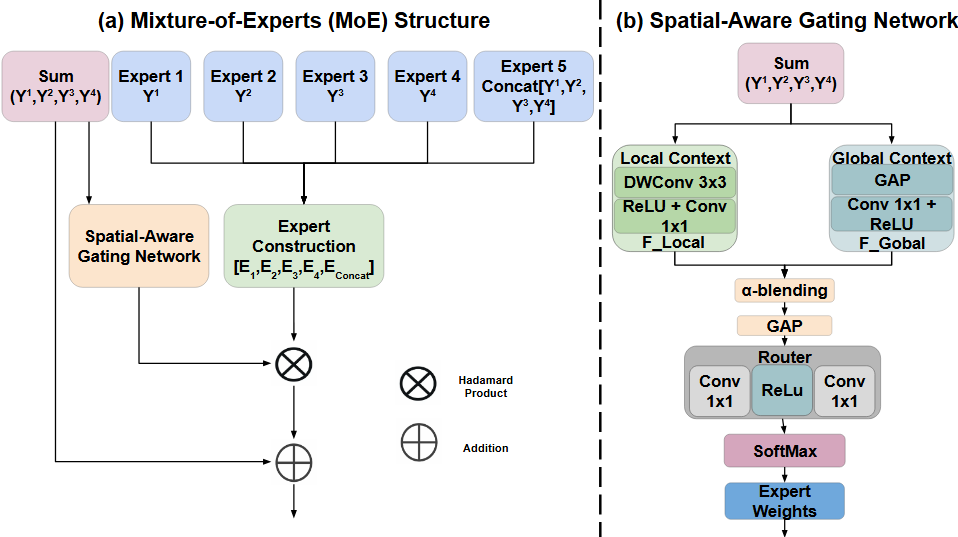}
%\vspace{-0.35cm}
\caption{Overview of the proposed MoE-based directional fusion module.}
%\vspace{-2mm}
\label{fig:moe}
\end{figure}
%\vspace{-2mm}
%%%%%%%%%%%%%%%%%%%%%%%%%%%%%%%%%%%%%%
\subsection{MoE-Based Directional Fusion}

\label{sec:moe}
In existing Mamba-based segmentation networks~\cite{adame2025topo,ruan2024vm,zhang2024vm}, feature fusion across scan directions typically relies on simple summation of the directional outputs. This fixed aggregation strategy is agnostic to the underlying object scale and local image complexity. It implicitly assumes that all directions and receptive fields are equally informative at every spatial location, which is rarely true in practice. Regions containing small, subtle lesions may require fine-scale, locally focused modeling, whereas larger structures and cluttered backgrounds may benefit more from broader, long-range context. As a consequence, such hand-crafted fusion can lead to inaccurate boundary localization and incomplete object masks, and it limits the model’s ability to adapt its representational capacity to different lesion sizes, shapes, and appearances.

\paragraph{Expert construction.}
To overcome these limitations, we treat the outputs of the directional scanners as a set of experts and introduce an MoE-based fusion module (Fig.~\ref{fig:moe}). Let $\{Y_l^{(1)}, \dots, Y_l^{(4)}\}$ denote the four directional feature maps at stage $l$, each in $\mathbb{R}^{C_l \times H_l \times W_l}$. We first normalize each map with Group Normalization to stabilize expert routing and reduce scale discrepancies across directions:
$
\widehat{Y}_l^{(i)} = \text{GN}(Y_l^{(i)}), \quad i = 1,\dots,4.
$
These four normalized maps form the first four experts:
$
E_1 = \widehat{Y}_l^{(1)},\; E_2 = \widehat{Y}_l^{(2)},\; E_3 = \widehat{Y}_l^{(3)},\; E_4 = \widehat{Y}_l^{(4)}.
$
In addition to these direction-specific experts, we construct a fifth \emph{concatenation expert} that explicitly models interactions among all directions. We concatenate the four normalized feature maps along the channel dimension and project them back to $C_l$ channels using a $1 \times 1$ convolution followed by batch normalization and ReLU:
$
E_{\text{concat}} = \phi\Big(\text{Conv}_{1\times1}\big(\text{Concat}(\widehat{Y}_l^{(1)},\dots,\widehat{Y}_l^{(4)})\big)\Big),
$
where $\phi$ denotes batch normalization and ReLU activation. This expert captures cross-directional correlations that cannot be represented by any single directional map alone, providing a more holistic, direction-agnostic view that is especially useful when lesion appearance is not aligned with a particular scanning direction. The final expert stack is
$
\mathcal{E}_l = [E_1, E_2, E_3, E_4, E_{\text{concat}}].
$

\paragraph{Spatial-aware gating network.}
To adaptively fuse these experts, we design a spatial-aware gating network (router) that produces spatially varying expert weights (left block in Fig.~\ref{fig:moe}). The router is driven by both local and global context.
We first obtain a local context descriptor $F_{\text{local}}$ by summing the raw directional outputs and applying a depthwise $3 \times 3$ convolution:
$
F_{\text{local}} = \text{DWConv}_{3\times3}\Big(\sum_{i=1}^{4} Y_l^{(i)}\Big).
$
In parallel, we compute a global context descriptor $F_{\text{global}}$ via adaptive average pooling to a $1 \times 1$ spatial resolution:
$
F_{\text{global}} = \text{GAP}\Big(\sum_{i=1}^{4} Y_l^{(i)}\Big),
$
and broadcast it back to the spatial size $H_l \times W_l$. A learnable scalar $\alpha \in (0,1)$ then blends these two signals:
$
F_l = \alpha \cdot F_{\text{local}} + (1 - \alpha) \cdot F_{\text{global}}.
$
This $\alpha$-blending allows the router to smoothly trade off between fine local details and coarse global semantics, depending on what is more informative for a given dataset or training stage.
A two-layer $1 \times 1$ convolutional router maps $F_l$ to unnormalized logits for the five experts at each spatial location:
$
\mathbf{z}_l = \text{Router}(F_l) \in \mathbb{R}^{5 \times H_l \times W_l}.
$
Applying a softmax over the expert dimension yields spatially conditioned expert weights:
$
\mathbf{w}_l = \text{Softmax}(\mathbf{z}_l), \quad
\sum_{e=1}^{5} w_{l,e}(h,w) = 1,\;\forall (h,w).
$
The routed fusion is then computed as a weighted sum of experts:
$
\widetilde{Y}_l = \sum_{e=1}^{5} w_{l,e} \odot E_e,
$
where $\odot$ denotes element-wise multiplication and $e \in \{1,2,3,4,\text{concat}\}$ indexes the experts.

\paragraph{Residual stabilization.}
To prevent routing degeneracy and preserve the strong directional signals, we introduce a residual bypass that adds back the raw directional outputs:
$
Z_l = \widetilde{Y}_l + \sum_{i=1}^{4} Y_l^{(i)}.
$
This residual path ensures that the encoder retains a robust baseline response even when the gating network is not yet well trained, while the MoE focuses on refining the relative importance of each direction and the concatenation expert. The fused output $Z_l$ is then forwarded to the next encoder stage and/or skip-connection path, providing the SDI decoder with a feature representation that is spatially coherent, directionally enriched, and adaptively fused according to lesion scale and boundary complexity.

\paragraph{Patch-MoE VSS block.}
Combining the patch-ordered scanning in Section~\ref{sec:patch-scan} with the MoE-based fusion in Section~\ref{sec:moe}, we obtain the Patch-MoE VSS block that replaces the original VSS block in VM-UNetV2~\cite{zhang2024vm}. Given an input feature map $X_l \in \mathbb{R}^{C_l \times H_l \times W_l}$, we first apply four directional patch-ordered scans (forward, reverse, WH forward, WH reverse), each with its associated patch size $p^{(k_d)}$, to generate four 1D sequences. Each sequence is processed by a shared-architecture Mamba layer (state space model) and then reshaped back to the 2D grid, yielding the directional feature maps $\{Y_l^{(1)},\dots,Y_l^{(4)}\}$. These maps are passed through the expert construction block to form the expert stack $\mathcal{E}_l = [E_1, E_2, E_3, E_4, E_{\text{concat}}]$, and the spatial-aware gating network computes the routing weights that produce the routed fusion $\widetilde{Y}_l$. Finally, we add the residual sum of the raw directional outputs to obtain the block output
$
Z_l = \widetilde{Y}_l + \sum_{i=1}^{4} Y_l^{(i)},
$
which serves as the output of the Patch-MoE VSS block and is forwarded to the next encoder stage or the skip-connection path. In this way, the Patch-MoE VSS block preserves the overall VSS design while replacing fixed summation with adaptive, context-aware expert fusion.

\section{Experiments and Results} \label{exp}
\label{sec:exp}

% %%%%%%%%%%%%%%%%%%%%%%%%%%%%%%%%%%%%%%
%\setlength{\tabcolsep}{3pt}
\begin{table}[t]
\centering
\caption{Quantitative comparison of different methods on five polyp segmentation datasets. The best results are highlighted in bold; the same in the other tables. Each experiment is run 5 times with different random seeds.}%, and the mean and standard deviation across runs are reported.}
 \scalebox{0.67}{
\begin{tabular}{l|llll}
\hline
Datasets & Methods & DSC (\%) $\uparrow$  & IoU (\%) $\uparrow$ & MAE $\downarrow$   \\
\hline 
\multirow{6}{*}{Kvasir-SEG} & 
U-Net~\cite{ronneberger2015u} &83.19 $\pm$ 0.82	&75.58 $\pm$ 1.04 &0.0472 $\pm$ 0.0024	  \\
% &SANet~\cite{wei2021shallow}&85.96 $\pm$ 0.54	 &77.24 $\pm$ 0.60   &0.0384	$\pm$ 0.0022  \\
% & U-Net v2~\cite{peng2025u} &84.39 $\pm$ 6.09 &76.41 $\pm$ 7.56 &0.0363 $\pm$ 0.0018  \\
& U-Net v2~\cite{peng2025u} &90.84 $\pm$ 0.15 &85.31 $\pm$ 0.37 &0.0265 $\pm$ 0.0010  \\
& VM-UNet~\cite{ruan2024vm} &89.90 $\pm$ 0.98 &84.00 $\pm$ 1.41 &0.0294 $\pm$ 0.0034  \\
& VM-UNetV2~\cite{zhang2024vm} &90.82 $\pm$ 0.25 &85.30 $\pm$ 0.29 &0.0260 $\pm$ 0.0020
\\\cline{2-5}
& Patch-MoE Mamba (Ours)&\textbf{90.90 $\pm$ 0.35} &\textbf{85.32 $\pm$ 0.44} &\textbf{0.0258 $\pm$ 0.0015} \\
\hline 

\multirow{6}{*}{ClinicDB} & 
U-Net~\cite{ronneberger2015u} &85.68 $\pm$ 1.34 &79.58 $\pm$ 1.25 &0.0171 $\pm$ 0.0018	  \\
% &SANet~\cite{wei2021shallow}&83.20 $\pm$ 0.77	 &74.13 $\pm$ 0.80 &0.0232	$\pm$ 0.0009  \\
% & U-Net v2~\cite{peng2025u} &86.00 $\pm$ 0.79 &79.09 $\pm$ 1.10 &0.0168 $\pm$ 0.0028  \\
& U-Net v2~\cite{peng2025u} &89.67 $\pm$ 0.96 &84.38 $\pm$ 0.96 &0.0161 $\pm$ 0.0023  \\
& VM-UNet~\cite{ruan2024vm} &88.46 $\pm$ 0.89 &82.74 $\pm$ 0.92 &0.0156 $\pm$ 0.0032  \\
& VM-UNetV2~\cite{zhang2024vm} &90.52 $\pm$ 0.79 &85.34 $\pm$ 0.68 &0.0120 $\pm$ 0.0028
\\\cline{2-5}
& Patch-MoE Mamba (Ours)&\textbf{91.32 $\pm$ 0.39} &\textbf{86.05 $\pm$ 0.47} &\textbf{0.0104 $\pm$ 0.0011} \\
\hline 

\multirow{6}{*}{ColonDB} & 
U-Net~\cite{ronneberger2015u} &62.48 $\pm$ 2.36 &53.99 $\pm$ 2.34 &0.0488 $\pm$ 0.0025	  \\
% &SANet~\cite{wei2021shallow}&70.29 $\pm$ 1.36	 &60.02 $\pm$ 1.26   &0.0414 $\pm$ 0.0024	  \\
% & U-Net v2~\cite{peng2025u} &72.40 $\pm$ 1.08 &62.42 $\pm$ 1.09 &0.0430 $\pm$ 0.0035  \\
& U-Net v2~\cite{peng2025u} &76.23 $\pm$ 1.21 &69.00 $\pm$ 1.48 &0.0336 $\pm$ 0.0011  \\
& VM-UNet~\cite{ruan2024vm} &75.40 $\pm$ 1.01 &67.80 $\pm$ 1.16 &0.0356 $\pm$ 0.0020  \\
& VM-UNetV2~\cite{zhang2024vm} &76.62 $\pm$ 1.28 &68.64 $\pm$ 1.02 &0.0336 $\pm$ 0.0022
\\\cline{2-5}
& Patch-MoE Mamba (Ours)&\textbf{77.94 $\pm$ 1.60} &\textbf{69.65 $\pm$ 1.31} &\textbf{0.0314 $\pm$ 0.0019} \\
\hline 

\multirow{6}{*}{ETIS} & 
U-Net~\cite{ronneberger2015u} &42.82 $\pm$ 1.38 &36.36 $\pm$ 1.22 &0.0365 $\pm$ 0.0039	  \\
% &SANet~\cite{wei2021shallow}&59.78 $\pm$ 0.60	 &49.58 $\pm$ 0.75  &0.0244 $\pm$ 0.0021	  \\
% & U-Net v2~\cite{peng2025u} &61.43 $\pm$ 2.12 &51.70 $\pm$ 2.24 &0.0382 $\pm$ 0.0072  \\
& U-Net v2~\cite{peng2025u} &72.83 $\pm$ 1.70 &63.78 $\pm$ 2.01 &0.0242 $\pm$ 0.0021  \\
& VM-UNet~\cite{ruan2024vm} &70.56 $\pm$ 2.27 &62.40 $\pm$ 2.19 &0.0240 $\pm$ 0.0094  \\
& VM-UNetV2~\cite{zhang2024vm} &72.56 $\pm$ 1.55 &63.68 $\pm$ 1.65 &\textbf{0.0180 $\pm$ 0.0016}
\\\cline{2-5}
& Patch-MoE Mamba (Ours)&\textbf{74.04 $\pm$ 0.78} &\textbf{64.86 $\pm$ 0.94} &{0.0196 $\pm$ 0.0006} \\
\hline 

\multirow{6}{*}{CVC-300} & 
U-Net~\cite{ronneberger2015u} &77.67 $\pm$ 2.00 &69.70 $\pm$ 1.63 &0.0149 $\pm$ 0.0016	  \\
% &SANet~\cite{wei2021shallow}&\textbf{87.58 $\pm$ 0.86}	 &\textbf{80.75 $\pm$ 1.01}   &0.0090 $\pm$ 0.0003	  \\
% & U-Net v2~\cite{peng2025u} &83.46 $\pm$ 1.16 &74.99 $\pm$ 1.08 &0.0128 $\pm$ 0.0019  \\
& U-Net v2~\cite{peng2025u} &85.88 $\pm$ 1.46 &78.99 $\pm$ 1.53 &0.0111 $\pm$ 0.0012  \\
& VM-UNet~\cite{ruan2024vm} &86.72 $\pm$ 1.25 &\textbf{80.42 $\pm$ 1.32} &0.0098 $\pm$ 0.0029  \\
& VM-UNetV2~\cite{zhang2024vm} &86.80 $\pm$ 1.21 &79.50 $\pm$ 1.10 &0.0086 $\pm$ 0.0015
\\\cline{2-5}
& Patch-MoE Mamba (Ours)&\textbf{87.31 $\pm$ 0.42} &{79.91 $\pm$ 0.77} &\textbf{0.0078 $\pm$ 0.0008} \\
\hline 
\end{tabular}
}
\label{tab:polyp}
\end{table}
%\vspace{-2mm}
% %%%%%%%%%%%%%%%%%%%%%%%%%%%%%%%%%%%%%%

%%%%%%%%%%%%%%%%%%%%%%%%%%%%%%%%%%%%%%
%\setlength{\tabcolsep}{3pt}
\begin{table}[t]
\centering
\caption{Quantitative comparison of different methods on ISIC 2017 and ISIC 2018 datasets.}
 \scalebox{0.7}{
\begin{tabular}{l|llll}
\hline
Datasets & Methods & DSC (\%) $\uparrow$  & IoU (\%) $\uparrow$ & MAE    \\
\hline 
\multirow{6}{*}{ISIC 2017}  
& U-Net~\cite{ronneberger2015u} 
& $87.48 \pm 0.19$ 
& $79.89 \pm 0.24$ 
& $0.0413 \pm 0.0011$ \\
% &SANet~\cite{wei2021shallow}
% & $87.94 \pm 1.85$	 
% & $79.97 \pm 2.86$  
% & $0.0370 \pm 0.0041$	  \\
% & U-Net v2~\cite{peng2025u}
% & $87.69 \pm 0.19$
% & $79.94 \pm 0.63$
% & $0.0383 \pm 0.0018$ \\
& U-Net v2~\cite{peng2025u}
& $87.90 \pm 0.34$
& $80.02 \pm 0.39$
& $0.0368 \pm 0.0016$ \\
& VM-UNet~\cite{ruan2024vm}
& $90.03 \pm 0.11$
& $83.35 \pm 0.24$
& $0.0317 \pm 0.0009$ \\
& VM-UNetV2~\cite{zhang2024vm} 
& $90.23 \pm 0.77$	 
& $83.59 \pm 0.91$  
& $0.0310 \pm 0.0023$
\\\cline{2-5}
& Patch-MoE Mamba (Ours)
& $\mathbf{90.85 \pm 0.93}$
& $\mathbf{84.45 \pm 1.23}$
& $\mathbf{0.0293 \pm 0.0026}$
\\\hline 

\multirow{6}{*}{ISIC 2018} 
& U-Net~\cite{ronneberger2015u}
& $86.82 \pm 0.33$
& $78.64 \pm 0.38$
& $0.0660 \pm 0.0025$ \\
% & SANet~\cite{wei2021shallow}         
% & $86.32 \pm 0.31$   
% & $77.50 \pm 0.35$  
% & $0.0594 \pm 0.0015$  \\
% & U-Net v2~\cite{peng2025u}     
% & $88.14 \pm 0.40$
% & $80.24 \pm 0.52$
% & $0.0528 \pm 0.0016$ \\
& U-Net v2~\cite{peng2025u}     
& $88.51 \pm 0.43$
& $81.06 \pm 0.48$
& $0.0524 \pm 0.0039$ \\
& VM-UNet~\cite{ruan2024vm}      
& $87.42 \pm 0.47$
& $79.66 \pm 0.65$
& $0.0558 \pm 0.0017$ \\
& VM-UNetV2~\cite{zhang2024vm}
& $88.36 \pm 0.33$
& $80.90 \pm 0.40$
& $0.0550 \pm 0.0021$ \\
\cline{2-5}
& Patch-MoE Mamba (Ours)
& $\mathbf{89.34 \pm 0.39}$
& $\mathbf{82.28 \pm 0.58}$
& $\mathbf{0.0496 \pm 0.0023}$ \\
\hline 
\end{tabular}
}
\label{tab:isic}
\end{table}
%\vspace{-2mm}
%%%%%%%%%%%%%%%%%%%%%%%%%%%%%%%%%%%%%%

%%%%%%%%%%%%%%%%%%%%%%%%%%%%%%%%%%%%%%%
% Requires: \usepackage{booktabs}
\begin{table}[t]
\centering
%\vspace{-2mm}
\caption{Ablation study on different components of Patch-MoE Mamba.}
%\vspace{-2mm}
\scalebox{0.67}{
\begin{tabular}{lcccccc}
\toprule
Method & Kvasir-SEG & ClinicDB & ColonDB & ETIS & CVC-300 & Average  \\
\midrule

\makecell{VM-UNetV2~\cite{zhang2024vm}}
& 90.82 & 90.52 & 76.62 & 72.56 & 86.80 & 83.46 \\\hline

\makecell{VM-UNetV2 \\ w/ Patch-Ordered Scanning}
& \textbf{91.14} & 91.12 & 76.68 & 73.76 & \textbf{87.40} & 84.02 \\\hline

\makecell{VM-UNetV2 \\ w/ Patch-Ordered Scanning\\ + MoE Fusion (Ours)}
& 90.90 & \textbf{91.32} & \textbf{77.94} & \textbf{74.04} & 87.31 & \textbf{84.30} \\
\bottomrule
\end{tabular}
}
\label{tab:ablation1}
\end{table}
%\vspace{-2mm}
%%%%%%%%%%%%%%%%%%%%%%%%%%%%%%%%%%%%%%%

% %%%%%%%%%%%%%%%%%%%%%%%%%%%%%%%%%%%%%%
% \begin{table}[t]
% \centering
% \caption{Ablation study on the components of the MoE-based directional fusion module.}
% %\vspace{-2mm}
% \scalebox{0.6}{
% \begin{tabular}{lcccccc}
% \toprule
% Method & Kvasir-SEG & ClinicDB & ColonDB & ETIS & CVC-300 & Average  \\
% \midrule

% Patch-MoE w/o Concat Expert  
% & 90.70 & 90.80 & 76.90 & 73.50 & 86.00 & 83.58 \\\hline

% Patch-MoE w/o Residual Addition  
% & \textbf{90.90} & 90.90 & 77.10 & 73.10 & 86.60 & 83.72 \\\hline

% Patch-MoE Mamba (Ours)
% & \textbf{90.90} & \textbf{91.32} & \textbf{77.94} & \textbf{74.04} & \textbf{87.31} & \textbf{84.30} \\

% \bottomrule
% \end{tabular}
% }
% \label{tab:ablation_moe}
% \end{table}
% %\vspace{-2mm}
% %%%%%%%%%%%%%%%%%%%%%%%%%%%%%%%%%%%%%%

%%%%%%%%%%%%%%%%%%%%%%%%%%%%%%%%%%%%%%
% Requires: \usepackage{booktabs}
\begin{table}[t]
\centering
\caption{Results for different patch sizes at different stages.}
%\vspace{-2mm}
\scalebox{0.74}{
\begin{tabular}{lcccccc}
\toprule
Patch Sizes & Kvasir-SEG & ClinicDB & ColonDB & ETIS & CVC-300 & Average  \\
\midrule
8844/8844/8844/8844  & 90.70 & 90.00 & 75.67 & 71.58 & 86.41 & 82.87 \\\hline
8844/8844/8844/1111  & 90.61 & 90.02 & 75.97 & 71.43 & 85.60 & 82.73 \\\hline
8844/8844/1111/1111  & 90.47 & 89.95 & 76.32 & 71.55 & 86.54 & 82.97 \\\hline
8844/1111/1111/1111  & \textbf{91.13} & \textbf{91.10} & \textbf{76.69} & \textbf{73.74} & \textbf{87.41} & \textbf{84.01} \\
\bottomrule
\end{tabular}
}
\label{tab:ablation2}
\end{table}
\subsection{Datasets and Experimental Setup}
\textbf{Polyp segmentation.}
We conduct experiments on five public polyp segmentation datasets: Kvasir-SEG~\cite{jha2020kvasir}, ClinicDB~\cite{bernal2015wm}, ColonDB~\cite{tajbakhsh2015automated}, ETIS~\cite{silva2014toward}, and CVC-300~\cite{vazquez2017benchmark}. For fair comparison, we follow the protocol in~\cite{peng2025u}: the training set uses 900 images from Kvasir-SEG and 550 from ClinicDB, and the test set includes CVC-300 (60), ColonDB (380), ETIS (196), plus 100 images from Kvasir-SEG and 62 from ClinicDB.
\textbf{Skin lesion segmentation.}
We further evaluate on ISIC 2017~\cite{codella2018skin} (2{,}150 images) and ISIC 2018~\cite{codella2019skin,tschandl2018ham10000} (2{,}694 images), using the train/test splits in~\cite{peng2025u}.
All experiments use PyTorch. Models are trained on an NVIDIA Tesla A100 (80\,GB) with AdamW (lr $1\times10^{-3}$, batch size 80). Following~\cite{zhang2024vm}, images are resized to $256\times256$. We use cosine annealing (min lr $1\times10^{-5}$, cycle length 50) for 300 epochs. The Vision Mamba encoder in VM-UNetV2 is initialized with pretrained VMamba-S~\cite{liu2024vmamba}. Standard augmentations (random flipping/rotation) are applied.

\subsection{Experimental Results}
Table~\ref{tab:polyp} compares Patch-MoE Mamba with U-Net, U-Net v2, VM-UNet, and VM-UNetV2 on five polyp datasets. Patch-MoE Mamba achieves the best Dice on all datasets, with competitive IoU and MAE, indicating that patch-ordered scanning with adaptive MoE fusion consistently improves performance across colonoscopic benchmarks. Notably, the largest gain over the prior SOTA VM-UNetV2 occurs on the most challenging ETIS set: Dice increases from 72.83 to 74.04 (+1.21). ETIS contains low-contrast images, small polyps, and irregular boundaries, so this improvement highlights the benefit of preserving local spatial structure for better spatial modeling and boundary localization.
% Table~\ref{tab:polyp} presents a quantitative comparison of Patch-MoE Mamba against U-Net, U-Net v2, VM-UNet, and VM-UNetV2 on the five public polyp segmentation datasets. From these results, we make the following observations. First, Patch-MoE Mamba achieves the best performance on all five datasets in terms of Dice score, while also producing highly competitive IoU and MAE values. This demonstrates that the proposed patch-ordered scanning and adaptive MoE fusion lead to consistent performance gains across diverse colonoscopic benchmarks. 
% More importantly, the largest improvement over the previous SOTA model VM-UNetV2 is observed on the most challenging dataset, ETIS. Specifically, Patch-MoE Mamba improves the Dice score from 72.56 to 74.04, corresponding to an absolute gain of 1.48 Dice points. This dataset contains low-contrast images, small-scale polyps, and highly irregular lesion boundaries, making it particularly sensitive to spatial modeling and boundary localization. The strong improvement on ETIS verifies the effectiveness of preserving local spatial structure via patch-ordered sequence construction. %and adaptively weighting multi-directional scan responses through MoE fusion.

\begin{figure}[h!]
\centering
\includegraphics[width=.4\textwidth]{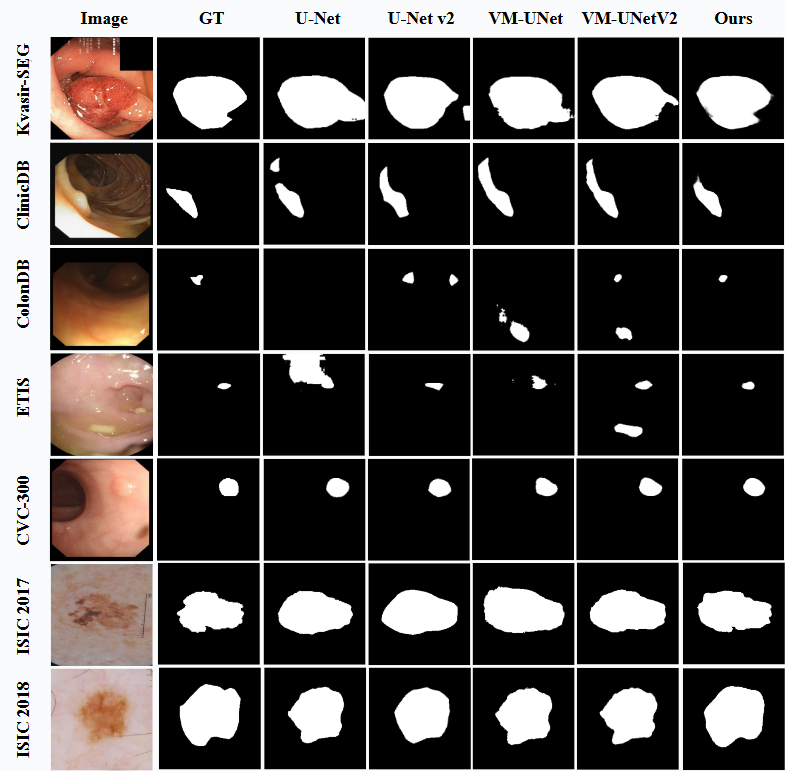}
\caption{Visual examples of segmentations results.}
\label{fig:visual-results}
\end{figure}

Table~\ref{tab:isic} reports results on ISIC 2017/2018. Patch-MoE Mamba achieves the best performance on both datasets, indicating strong generalization from colonoscopy images to skin lesion segmentation with different visual characteristics.
Fig.~\ref{fig:visual-results} shows qualitative examples across the seven datasets. Compared with baselines, Patch-MoE Mamba produces cleaner masks with fewer false positives and sharper boundaries. Notably, it better suppresses background spurious activations on ColonDB/ETIS and delineates polyp boundaries more accurately on Kvasir-SEG.%, yielding predictions closer to the GT.

% Table~\ref{tab:isic} further reports performance on the ISIC 2017 and ISIC 2018 skin lesion segmentation datasets. Patch-MoE Mamba achieves the best results on both benchmarks, confirming that the proposed architecture generalizes beyond colonoscopic imagery to a different medical imaging modality with distinct appearance characteristics.

% Fig.~\ref{fig:visual-results} illustrates several qualitative segmentation examples from the seven datasets. From these visual results, we observe that Patch-MoE Mamba produces segmentation masks with cleaner boundaries and fewer false positive predictions compared with the baseline methods. In particular, our method more effectively suppresses spurious activations in background regions (as shown on ColonDB and ETIS) and produces sharper boundary delineation (as shown on Kvasir-SEG), yielding predictions that are closer to the GT.

%%%%%%%%%%%%%%%%%%%%%%%%%%%%%%%%%%%%%%
%\vspace{-2mm}
%%%%%%%%%%%%%%%%%%%%%%%%%%%%%%%%%%%%%%
\subsection{Ablation Study}
Table~\ref{tab:ablation1} analyzes the effects of patch-ordered scanning and the MoE fusion module. From the VM-UNetV2 baseline, patch-ordered scanning alone improves the average Dice from 83.46 to 84.02, indicating that patch-wise, spatially coherent token ordering strengthens feature learning. Adding the MoE fusion module further boosts performance to 84.30, validating the proposed MoE-based directional fusion.
Table~\ref{tab:ablation2} studies patch-size configurations. The setting $8844/1111/1111/1111$ (stages 1--4) achieves the best average Dice of 84.01, suggesting that coarse scanning in early stages and finer scanning in deeper stages better captures multi-scale context while preserving boundary details.

\begin{table}[t]
\centering
\caption{Comparison of computational complexity of different models.}
\label{tab:complexity_moe}
%\vspace{0.05cm}
\scalebox{0.7}{
\begin{tabular}{cccc}
\hline
{Model} & {Input size} & {Params (M) $\downarrow$} & {FLOPs (G) $\downarrow$}  \\ \hline

\makecell{U-Net v2~\cite{peng2025u}}          
& (3, 256, 256)       
& 25.15        
& 5.58            
\\\hline

\makecell{VM-UNetV2~\cite{zhang2024vm}}          
& (3, 256, 256)       
& \textbf{22.77}          
& \textbf{5.31}            
\\\hline

\makecell{Patch-MoE w/o Concat Expert}          
& (3, 256, 256)       
& 28.44          
& 10.03            
\\\hline

\makecell{Patch-MoE w/o Residual Addition}          
& (3, 256, 256)       
& 70.06          
& 28.18            
\\\hline

\makecell{Patch-MoE Mamba (Ours)}          
& (3, 256, 256)       
& 70.06          
& 28.18            
\\\hline

\end{tabular}
}
\end{table}
%%%%%%%%%%%%%%%%%%%%%%%%%%%%%%%%%%%%%%
% \subsection{Computational Complexity}
% We report the number of parameters and FLOPs for all models.
% As shown in Table~\ref{tab:complexity_moe}, the introduction of patch-ordered scanning does not change the overall computational order, as it only reorganizes the token traversal sequence while preserving the total number of tokens. 
% In contrast, the concatenation expert substantially increases the computational cost. When the concatenation expert is enabled, the parameter count rises to 70.06M and the FLOPs increase to 28.18 GFLOPs. This is expected, as the concatenation expert explicitly aggregates the four directional feature maps along the channel dimension and applies an additional $1\times1$ projection layer, leading to denser channel interactions and higher computational overhead.
\subsection{Computational Complexity}
We report parameters and FLOPs for all models. As shown in Table~\ref{tab:complexity_moe}, patch-ordered scanning preserves the overall computational order because it only reorders token traversal while keeping the token count unchanged. In contrast, the concatenation expert markedly increases cost: enabling it raises parameters to 70.06M and FLOPs to 28.18\,GFLOPs. This overhead is expected, since it concatenates four directional feature maps along channels and adds an extra $1\times1$ projection, resulting in denser channel interactions.

\section{Conclusions}
We presented \textit{Patch-MoE Mamba}, a patch-ordered mixture-of-experts state space architecture for medical image segmentation. The proposed method combines hierarchical patch-ordered scanning to preserve local spatial neighborhoods and capture multi-scale context with an MoE-based fusion module that adaptively integrates multiple directional Mamba outputs and a learnable concatenation expert. A residual directional summation further stabilizes fusion and preserves discriminative features. Experiments on five polyp segmentation datasets and two skin lesion segmentation datasets demonstrate the effectiveness and generality of Patch-MoE Mamba.

\section{Acknowledgements}
%\vspace{-2mm}
This research was supported in part by NSF grants CCF-2523787, 2112650, 2434916, ELE250047, and 2018900, and AHA award 26AIREA1574568.

\bibliographystyle{IEEEtran}
\bibliography{references}

\end{document}